\documentclass[11pt,letterpaper]{article}
\usepackage{ijcnlp2017}
\usepackage{times}
\usepackage{latexsym}
\usepackage{graphicx}
\usepackage{booktabs}
\usepackage{multirow}
\usepackage{subfig}
\usepackage{url}
\usepackage{tabularx}
\usepackage{footnote}
\makesavenoteenv{tabular}
\usepackage{makecell}

\ijcnlpfinalcopy

\title{Using Context Events in Neural Network Models for Event Temporal Status Identification}

\author{Zeyu Dai, Wenlin Yao, Ruihong Huang \\
	   Department of Computer Science and Engineering \\
       Texas A\&M University \\
  {\tt \{jzdaizeyu, wenlinyao, huangrh\}@tamu.edu}}

\date{}

\begin{document}
\maketitle
\begin{abstract}
Focusing on the task of identifying event temporal status, we find that events directly or indirectly governing the target event in a dependency tree are most important contexts. 
Therefore, we extract dependency chains containing context events and use them as input in neural network models, which consistently outperform previous models using local context words as input. 
Visualization verifies that the dependency chain representation can effectively capture the
context events which are closely related to the target event and play key roles in predicting event temporal status.
\end{abstract}

\section{Introduction}
Event temporal status identification aims to recognize whether an event has happened (PAST), is currently happening (ON-GOING) or has not happened yet (FUTURE), which can be crucial for event prediction, timeline generation and news summarization. 

Our prior work \cite{huang-EtAl:2016:EMNLP2016} showed that linguistic features, such as tense, aspect and time expressions, are insufficient for this challenging task, which instead requires accurate understanding of composite meanings from wide sentential contexts. 
However surprisingly, the best performance for event temporal status identification was achieved by a Convolutional Neural Network (CNN) running on local contexts (seven words to each side) surrounding the target event \cite{huang-EtAl:2016:EMNLP2016}.

Considering the following sentence with a future event ``protest'':

(1) {\it Climate activists from around the world will launch a hunger strike here on Friday, describing their 
\textbf{protest (FU)} as a ``moral reaction to an immoral situation'' in the face of environmental catastrophe.}

\begin{figure*}[h]
\includegraphics[scale=0.43]{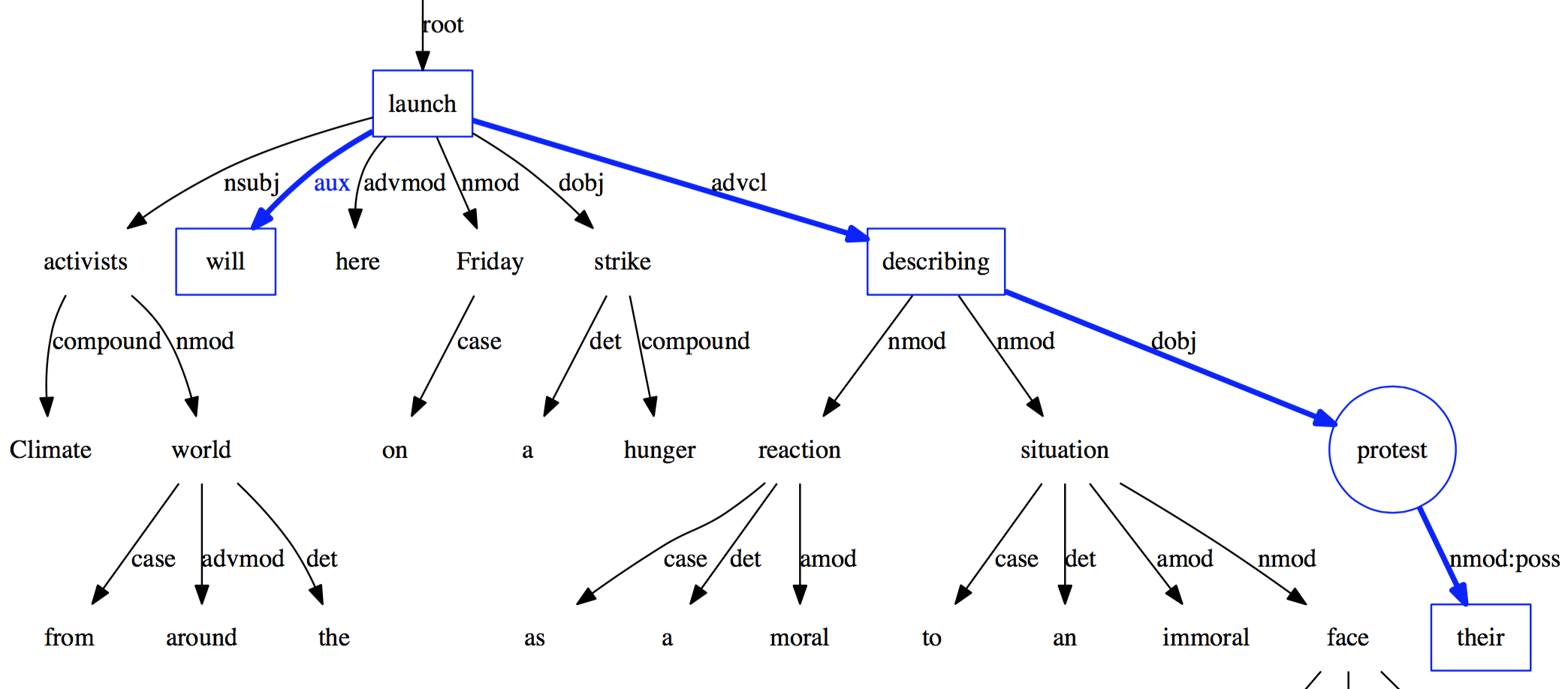}
\caption{The dependency parse tree for the example (1). The blue and bold edges show the extracted dependency chain for the target event ``protest'' (in circle).}
\label{dependency parsing chain}%
\end{figure*}

The local context ``describing'' may mislead the classifier that this is an on-going event,  while the actual future temporal status indicative words, ``will launch'', appear nine words away from the target event mention ``protest''. 
Clearly, the local window of contexts is filled with irrelevant words, meanwhile, it fails to capture important temporal status indicators.
However, as shown in figure \ref{dependency parsing chain}, the event ``launch'' is actually a high order indirect governor word of the target event word ``protest'' and is only two words away from the target event in the dependency tree.
Indeed, we observed that the temporal status indicators are often words that syntactically govern or depend on the event mention at all levels of a dependency tree. 
Furthermore, we observed that higher level governing words in dependency trees frequently refer to events as well, which are closely related to the target event and useful to predict its temporal status.

Following these observations, we form a dependency chain of relevant contexts by extracting words that appear between an event mention and the root of a dependency parse tree as well as words that are governed by the event mention. 
Then we use the extracted dependency chain as input for neural network classifiers. 
This is an elegant method to capture long-distance dependency between events within a sentence.
It is known that a verb and its direct or indirect governor can be far away in a word sequence if modifiers such as adjectives or clauses lie in between, but they are adjacent in the parse tree.

Experimental results using two neural network models (i.e., LSTMs and CNNs) show that classifiers with dependency chains as input can better capture temporal status indicative contexts and clearly outperform the ones with local contexts.
Furthermore, the models with dependency chains as input outperform a tree-LSTM model in which the full dependency trees are used as input.
Visualization reveals that it is much easier for neural net models to identify and concentrate on the relevant regions of contexts using the dependency chain representation, which further 
verifies that additional event words in a sentential context are crucial to predict target event temporal status.

\section{Related Work}
Constituency-based and dependency-based parse trees have been explored and applied to improve performance of neural nets for the task of sentiment analysis and semantic relation extraction~\cite{socher-EtAl:2013:EMNLP,bowman2015recursive,Tai15improvedsemantic}. 
The focus of these prior studies is on designing new neural network architectures (e.g., tree-structured LSTMs) corresponding to the parse tree structure. In contrast, our method aims at extracting appropriate event-centered data representations from dependency trees so that the neural net models can effectively concentrate on relevant regions of contexts. 

Similar to our dependency chains, dependency paths between two nodes in a dependency tree have been widely used as features for various NLP tasks and applications, including relation extraction \cite{bunescu2005shortest}, temporal relation identification \cite{choubey2017sequential} semantic parsing \cite{moschitti2004study} and question answering \cite{cui2005question}. Differently, our dependency chains are generated with respect to an event word and include words that govern or depend on the event, which therefore are not bounded by two pre-identified nodes in a dependency tree.

\begin{table*}[h]
\centering
\begin{tabular}{|l|c|c|c|c|c|}
\hline
Model                  & PA & OG & FU & Macro & Micro \\ \hline
\multicolumn{6}{|c|}{Local Contexts} \\ \hline
CNN \cite{huang-EtAl:2016:EMNLP2016} & 91/83/87 & 46/57/51 & 49/67/57 & 62/69/65 & 77/77/76.9 \\
LSTM                           & 88/83/85 & 47/54/51 & 52/62/57 & 63/66/64 & 75/75/75.5\\ \hline
\multicolumn{6}{|c|}{Dependency Chains} \\ \hline     
CNN & 91/84/87 & 49/63/55 & 60/65/62 & 67/71/68 & 79/79/78.6\\ 
LSTM              & 92/85/\textbf{88} & 49/63/\textbf{55} & 63/71/\textbf{67} & 68/73/\textbf{70} & 80/80/\textbf{79.6}\\ \hline
\multicolumn{6}{|c|}{Full Dependency Trees} \\ \hline   
tree-LSTM                & 92/80/86   &47/59/53    &30/58/40    &56/66/60       &75/75/75.1       \\ \hline
\end{tabular}
\caption{Classification results on the test set. Each cell shows Recall/Precision/F1 score.}
\label{Classification Result on test set.}
\end{table*}

\section{Dependency Chain Extraction}
Figure \ref{dependency parsing chain} shows the dependency parse tree\footnote{We used the Stanford CoreNLP to generate dependency parse trees.} for the example (1). To extract the dependency chain for the target event, we have used a two-stage approach to create the chain. 
In the first stage, we start from the target event, traverse the dependency parse tree, identify all its direct or 
indirect governors and dependents and include these words in the chain. For the example (1), a list of words [launch, describing, protest, their] are included in the dependency chain after the first stage.

Then in the second stage, we apply one heuristic rule to extract extra words from the dependency tree.
Specifically, if a word is in a particular dependency relation\footnote{\url{http://nlp.stanford.edu/software/dependencies_manual.pdf}}, {\it aux}, {\it auxpass} or {\it cop}, with a word that is already in the chain after the first stage, then we include this word in the chain as well. 
For the example (1), the word ``will'' is inserted into the dependency chain in the second stage. 
The reason we perform this additional step is that context words identified with one of the above three dependency relations usually indicate tense, aspect or mood of a context event, which are useful in determining the target event's temporal status. 

For each extracted dependency chain, we sort the words in the chain according to their textual order in the original sentence. Then the ordered sequence of words will be used as the input for LSTMs and CNNs.

\section{Experiments}
\subsection{The EventStatus Corpus}
We experiment on the EventStatus corpus \cite{huang-EtAl:2016:EMNLP2016}, which contains 4500 English and Spanish news articles about civil unrest events (e.g., protest and march), where each civil unrest event mention has been annotated with three categories, {\bf Past (PA)}, {\bf On-Going (OG)} and {\bf Future (FU)}, to indicate if the event has concluded, is currently ongoing or has not happened yet. 

We only use the English portion of the corpus which include 2364 documents because our previous work \cite{huang-EtAl:2016:EMNLP2016} has reported that the quality of dependency parse trees generated for Spanish articles is poor and our approach heavily rely on dependency trees.  
Furthermore, we randomly split the data into tuning (20\%) and test (80\%) set, and conduct the final evaluation using 10-fold cross-validation on the test set, following the prior work \cite{huang-EtAl:2016:EMNLP2016}. Table \ref{Temporal Status Label Distribution} shows the distribution of each event temporal status category in the dataset. 
\begin{table}[ht]
\centering
\label{table 1}
\begin{tabular}{|c|c|c|c|}
\hline
                    & \textbf{PA} & \textbf{OG} & \textbf{FU} \\ \hline
\textbf{Test}   & 1406(67\%)  & 429(21\%)   & 254(12\%)   \\ \hline
\textbf{Tuning} & 354(61\%)   & 157(27\%)   & 66(12\%)    \\ \hline
\end{tabular}
\caption{Temporal Status Label Distribution}
\label{Temporal Status Label Distribution}
\end{table}

\subsection{Neural Network Models} 
In our experiments, we applied three types of neural network models including CNNs \cite{collobert2011natural,kim2014convolutional}, LSTMs \cite{schmidhuber1997long}, and tree-LSTMs \cite{Tai15improvedsemantic}. 
For CNNs, we used the same architecture and parameter settings as \cite{huang-EtAl:2016:EMNLP2016} with a filter size of 5. For LSTMs, we implemented a simple architecture that consists of one LSTM layer and one output layer with softmax function. For tree-LSTMs, we replicated the Dependency tree-LSTMs\footnote{Our tree-LSTMs implementation were adjusted from \url{ https://github.com/ttpro1995/TreeLSTMSentiment}} from \cite{Tai15improvedsemantic} and added an output layer on top of it. Both of the two latter neural nets used the same number (300) of hidden units as CNNs. Note that we have also experimented with complex LSTM models, including the ones with multiple layers, with bidirectional inferencing \cite{schuster1997bidirectional} as well as with attention mechanism \cite{bahdanau2014neural,DBLP:conf/emnlp/WangHZZ16}, however none of these complex models improve the event temporal status prediction performance.

All the models were trained using RMSProp optimizer with the initial learning rate 0.001 and the same random seed. 
In addition, we used the pre-trained 300-dimension English word2vec\footnote{Downloaded from \url{ https://docs.google.com/uc?id=0B7XkCwpI5KDYNlNUTTlSS21pQmM}} embeddings \cite{mikolov2013distributed,mikolov2013efficient}. 
The training epochs 
and dropout \cite{hinton2012improving} ratio 
for neural net layers were treated as hyperparameters and were tuned using the tuning set. 
The best LSTM model ran for 50 training epochs and used a dropout ratio of 0.5.

\begin{figure*}[h]
    \centering
    \subfloat[Local Contexts gives wrong prediction OG]{{\includegraphics[scale=0.37]{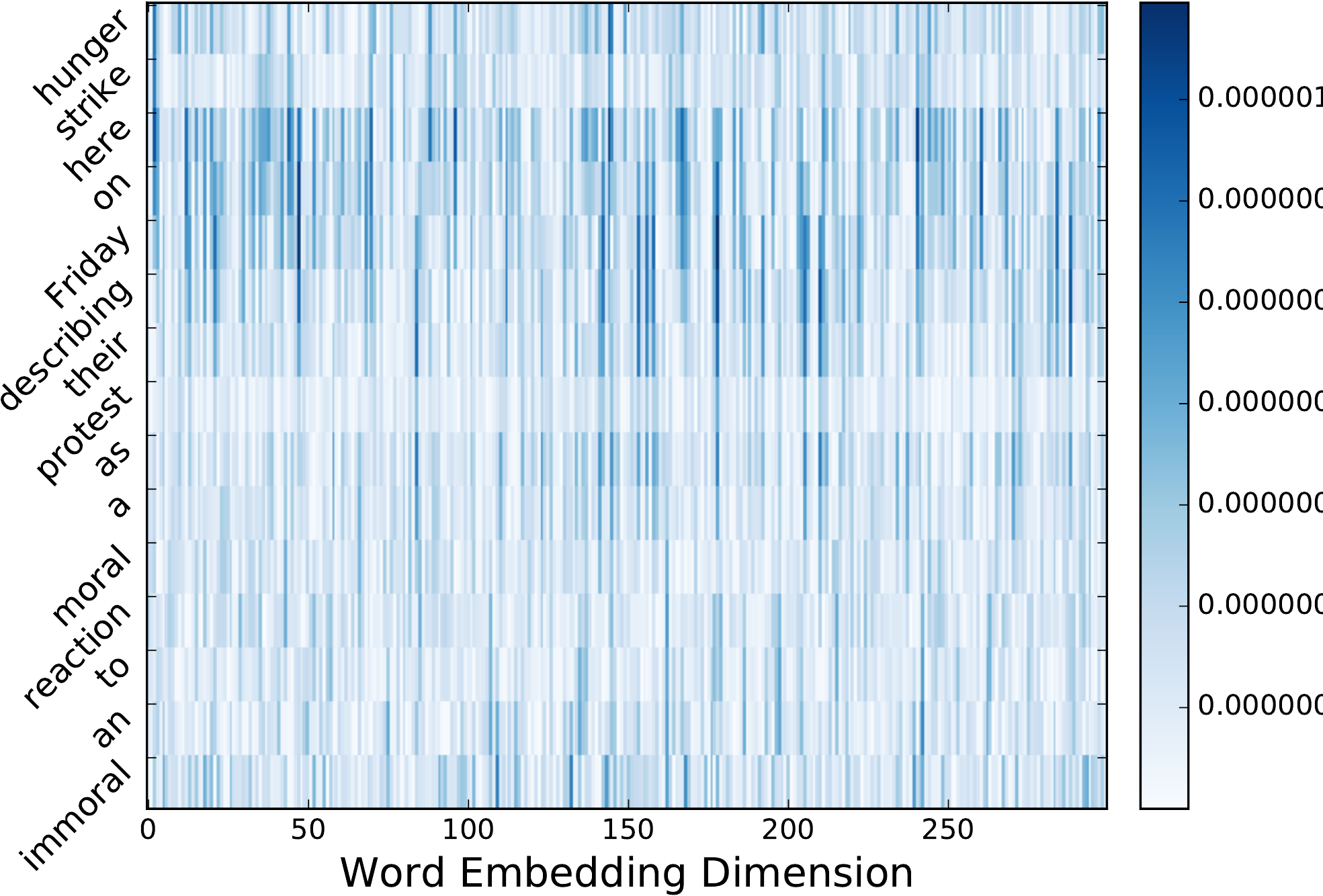} }}%
    \qquad
    \subfloat[Dependency Chains gives correct prediction FU]{{\includegraphics[scale=0.38
    ]{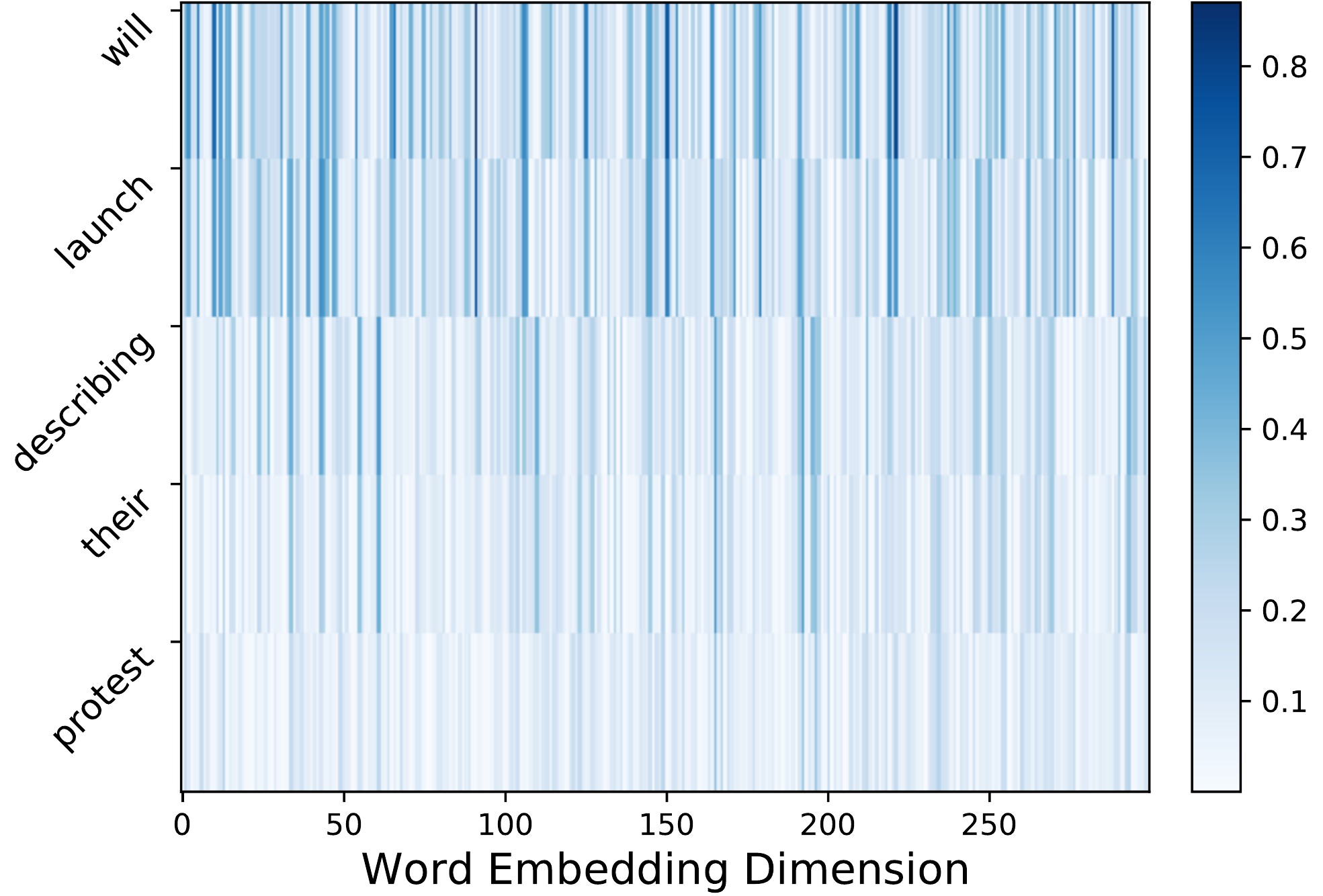} }}%
    \caption{Saliency heatmaps for the example (1). Local Context Input: {\it hunger strike here on Friday, describing their protest as a moral reaction to an immoral};  Dependency Chain Input: {\it will launch describing their protest}. A deeper color indicates that the corresponding word has a higher weight and contributes more to the prediction.}
    \label{heatmap2}%
\end{figure*}

\subsection{Classification Result}
Table \ref{Classification Result on test set.} shows the new experimental results on the test set as well as our prior results \cite{huang-EtAl:2016:EMNLP2016}.
For models using local contexts as input, 
the context window size
of 7 (7 words preceding and following the target event mention, therefore, 15 words in total) yields the best result, 
as reported in our prior paper \cite{huang-EtAl:2016:EMNLP2016}. 
Note that dependency chains we generated have an average length of 7.4 words in total, which are much shorter than 15 words of local contexts as used before.

From Table \ref{Classification Result on test set.}, we can see that both LSTM and CNN models using dependency chains as input consistently outperform the corresponding models using local contexts. 
Especially, the LSTM model running on dependency chains achieves the best performance of 70.0\% Macro and 79.6\% Micro F1-score, which outperforms the previous local context based CNN model \cite{huang-EtAl:2016:EMNLP2016} by a large margin. 
Statistical significance testing shows that the improvements are significant at the $p < 0.01$ level (t-test).
In particular for on-going and future events, the dependency chain based LSTM model improves the temporal status classification F-scores by 4 and 10 percentages respectively. 
In addition, the tree-LSTM model taking account of full dependency trees achieves a comparable result with local context based neural network models, but performs worse than dependency chain based models.
The reason why the tree-LSTM model does not work well is that irrelevant words, including adjective modifiers and irrelevant clauses forming branches of dependency trees, may distract the classifier and have negative effect in predicting the temporal status of an event. 

\section{Visualizing LSTM}
Following the approach used in \cite{li2016visualizing}, we drew salience heatmaps\footnote{Illustrate absolute values of derivatives of the loss function to each input dimension.} in order to understand contributions of individual words in a dependency chain to event temporal status identification. Figure \ref{heatmap2} shows heatmaps of LSTM models when applied to example (1) using its different data representations as input. 
We can clearly see that the dependency chain input effectively retains contexts that are relevant for predicting event temporal status.
Specifically, as shown in Figure \ref{heatmap2}(b), the context event ``launch'' that indirectly governs the target event ``protest'' in the dependency chain 
together with the auxiliary verb ``will'' have received the highest weights and are most useful in predicting the correct temporal status.

\section{Error Analysis}
More than 40\% of errors on the tuning set produced by our best LSTM model are due to the ``{\it Past or On-going} ambiguity'', which usually happen when there are few signals within a sentence that can indicate whether an event has concluded or not. 
In such scenarios, the classifier tends to predict the temporal status as {\it Past} since this event temporal status is majority in the dataset, which explains the low performance on predicting on-going events. 
To resolve such difficult cases, even wider contexts beyond one sentence should be considered. 

Around 10\% of errors are time expression related. The neural net models seem not be able to wisely use time expressions (e.g., ``in 1986'', ``on Wednesday'') without knowing the exact document creation time and their temporal relations.
In addition, some mislabelings occur because neural nets are unable to capture compositionality of multi-word expressions or phrasal verbs that alone can directly determine the temporal status of their following event, such as ``on eve of'' and ``go on''.

\section{Conclusion}
We presented a novel dependency chain based approach for event temporal status identification which can better capture relevant regions of contexts containing other events that directly or indirectly govern the target event. 
Experimental results showed that dependency chain based neural net models consistently outperform commonly used local context based models in predicting event temporal status, as well as a tree-structured neural network model taking account of complete dependency trees.
To further improve the performance of event temporal status identification, we believe that wider contexts beyond the current sentence containing an event should be exploited.

\section*{Acknowledgments}
We acknowledge the support of NVIDIA Corporation for their donation of one GeForce GTX TITAN X GPU used for this research.

\bibliography{ijcnlp2017}
\bibliographystyle{ijcnlp2017}

\end{document}